\documentclass[letterpaper, 10 pt, conference]{imitation_learning/format/ieeeconf} 
\IEEEoverridecommandlockouts       
\usepackage[normalem]{ulem}	                        
\usepackage[table,usenames,dvipsnames]{xcolor}      
\usepackage{extarrows}                              
\let\labelindent\relax
\usepackage{enumitem}                               
\let\proof\relax

\usepackage{amsmath,amsthm,amssymb,amsfonts,dsfont} 
\usepackage{cite}
\makeatletter
\newif\if@restonecol
\newif\foralgo@restonecol
\makeatother

\usepackage{graphicx}
\usepackage{subfigure}
\usepackage{tabularx}
\usepackage{booktabs}
\usepackage{multirow,multicol, array}
\usepackage[font={small}]{caption} 

\usepackage{algorithm,algorithmic}
\let\appendices\relax
\usepackage[titletoc,title]{appendix}
\usepackage{setspace}
\usepackage{latexsym}
\usepackage{bm}
\usepackage{mathrsfs}
\makeatletter
\let\NAT@parse\undefined
\makeatother
\usepackage[breaklinks=true, colorlinks, bookmarks=true, citecolor=Blue, urlcolor=Magenta,linkcolor=Black]{hyperref}

\usepackage{Definitions}

\newcommand{\dpi}{d}

\title{\LARGE \bf
Offline Imitation Learning upon Arbitrary Demonstrations by Pre-Training Dynamics Representations
}
\author{Haitong Ma, Bo Dai, Zhaolin Ren, Yebin Wang, Na Li
\thanks{Haitong Ma, Zhaolin Ren, Na Li are with School of Engineering and Applied Sciences, Harvard University. Bo Dai is with School of Computational Science and Engineering, Georgia Institute of Technology. Yebin Wang is with Mitsubishi Electric Research Laboratories (MERL) and Na Li is a visiting scholar at MERL.  Email: \texttt{\{haitongma, zhaolinren\}@g.harvard.edu, 
bodai@cc.gatech.edu, yebinwang@ieee.org,  nali@seas.harvard.edu}. 
}
\thanks{ The work is supported under NSF AI Institute: 2112085, NSF CNS: 2003111, NSF ECCS: 2401390, 2401391.
 }
\thanks{The Github link for open-sourced code, videos, and full report is available at \href{https://congharvard.github.io/repr-imitation-learning/}{https://congharvard.github.io/repr-imitation-learning/}.
}
}
\allowdisplaybreaks
\begin{document}
\maketitle
\begin{abstract}

Limited data has become a major bottleneck in scaling up offline imitation learning (IL).
In this paper, we propose enhancing IL performance under limited expert data by introducing a pre-training stage that learns \emph{dynamics representations}, derived from factorizations of the transition dynamics. We first theoretically justify that the optimal decision variable of offline IL lies in the representation space, significantly reducing the parameters to learn in the downstream IL. Moreover, the dynamics representations can be learned from arbitrary data collected with the same dynamics, allowing the reuse of massive non-expert data and mitigating the limited data issues. We present a tractable loss function inspired by noise contrastive estimation to learn the dynamics representations at the pre-training stage. 
Experiments on MuJoCo demonstrate that our proposed algorithm can mimic expert policies with as few as a single trajectory. Experiments on real quadrupeds show that we can leverage pre-trained dynamics representations from simulator data to learn to walk from a few real-world demonstrations.
    
\end{abstract}

\setlength{\abovedisplayskip}{2pt}
\setlength{\abovedisplayshortskip}{2pt}
\setlength{\belowdisplayskip}{2pt}
\setlength{\belowdisplayshortskip}{2pt}

\setlength{\floatsep}{1ex}
\setlength{\textfloatsep}{1ex}

\section{Introduction}

Offline imitation learning (IL) agents aim to mimic an expert policy  using only a fixed dataset of expert demonstrations, without interacting with the environment through a behavior policy. Since offline IL eliminates the cost and risks associated with trial-and-error exploration, it has been widely applied to various robot learning tasks, including manipulation~\cite{chi2023diffusion,mandlekar2021matters} and locomotion~\cite{ratliff2007imitation,peng2020learning}.

One of the most commonly used offline IL algorithms is behavior cloning \cite{pomerleau1988alvinn}, which uses supervised learning to match the expert policy and behavior policy directly, ignoring the Markovian properties. Behavior cloning suffers from the notorious compound error, meaning that a small learning error in a single step will drive the agent into scarce or unseen trajectories in the expert dataset, in turn amplifying the learning error after several steps~\cite{ross2011reduction}. 
To avoid compound error, 
IL is formulated as distribution matching between the behavior and expert state-action densities, leveraging the Markovian structure of expert data to mitigate the compound error~\cite{ho2016generative}. This approach has been further extended to the offline setting~\cite{kostrikov2019imitation} using DIstribution Correlation Estimation (DICE,~\cite{nachum2019dualdice}) as computation tools. 


The major challenges in solving distribution matching offline are twofold, \ie, limited expert data and solving $\min-\max$ optimization. 
In the offline setting, the agent can not interact with the environment and can only learn from limited expert data, resulting in heavy overfitting and poor generalization. Current solutions to limited data are mixing expert and sub-optimal data by regularization or weighted sum \cite{kostrikov2019imitation,kim2022demodice,ma2022versatile,sasaki2020behavioral}. However, the mixture still requires sub-optimality of the auxiliary data.
Another significant issue is the computational complexity of $\min$-$\max$ optimization~\cite{nachum2019dualdice, nachum2020reinforcement}, as DICE inherently performs a primal-dual optimization.  Specifically, with neural network parametrizations in practice, the $\min$-$\max$ optimization becomes highly unstable~\cite{arjovsky2017wasserstein}.  


To handle these challenges, we propose the \emph{dynamics representations} to further leverage the dynamics information to improve offline IL. 
Specifically, we define dynamics representations from the factorization of system dynamics and theoretically justify their abilities to represent 
the decision variables of offline IL optimization. 
Therefore, we can conduct offline IL on the representation space, mitigating the computational difficulties of solving $\min-\max$ with neural networks and reducing parameters to learn.
Moreover, the dynamics representations can be learned from arbitrary demonstrations with the same dynamics, relieving the limited data issue by learning representations from a large amount of non-expert or even random data. We formulate a two-stage algorithm that learn dynamic representations from all data with shared dynamics in the pre-training stage and conducts downstream IL on representation space with expert data only in the main stage. 

We validate the proposed algorithm through locomotion tasks in both the MuJoCo simulator and real quadrupeds. The MuJoCo experiments demonstrate that dynamics representations enable learning locomotion policies with as few as a single expert trajectory. Meanwhile, real-world quadruped experiments show that the agent can learn to walk using 1000 seconds of demonstration data collected from real hardware, built upon dynamics representations learned from simulators.



\section{Related works}

\subsection{Imitation Learning via Distribution Matching}

Distribution matching has online and offline variants depending on whether the agent can interact with the environment. 
\textbf{Online distribution matching} starts from inverse reinforcement learning (IRL), where the agent tries to recover the reward from expert trajectories \cite{ng2000algorithms,abbeel2004apprenticeship,syed2007game}. However, we need another RL process on the IRL output to recover the expert policy. \cite{ho2016generative} explains how the reward function learning can be bypassed to directly learn expert policy by distribution matching. The algorithm starts the adversarial IL family that adversarially trains a behavior policy to mimic the expert and a discriminator to discriminate behavior and expert trajectory \cite{xu2022discriminator,ghasemipour2020divergence}. \textbf{Offline distribution matching} is more challenging since we cannot sample from the behavior distribution, and the discriminator training is impossible. \cite{kostrikov2019imitation} first makes the offline distribution matching possible using DICE to estimate the density ratio between behavior policy and offline dataset~\cite{nachum2019dualdice}. Then offline IL focuses on the limited expert demonstration issue with different choices of regularization terms \cite{kostrikov2019imitation,kim2022demodice,kim2022lobsdice,ma2022versatile}.




\vspace{-1mm}
\subsection{Representation Learning for IL}
\vspace{-1mm}
In reinforcement learning (RL), successor representation is popular for learning and transfer \cite{dayan1993improving,barreto2017successor}, but it is defined over reward functions and not capable for IL. Some work re-parametrized policies using latent variable models to learn task-agnostic skills on the latent space to represent general knowledge \cite{yang2021trail,ajay2020opal}, but it is designed only for learning from experts.
In control theory, the Koopman operator theory \cite{brunton2021modern} tries to lift the problem to a higher dimensional space where it becomes a linear system, but learning such mappings is very difficult.  

Representations defined by factorization of transition dynamics \cite{jin_provably_2019,agarwal_flambe_2020} have recently gained significant interest due to their rich representational capacity and transferability. The representability of dynamics factorization has been justified by theoretical analyses \cite{zhang2022making,ren2022spectral} as well as empirical studies on sim-to-real transfer learning \cite{ma2024skill}. In this paper, we show that these representations are also compatible with IL after adding noise contrastive design inspired by contrastive learning~\cite{gutmann2012noise,zhang2022making}. 




    






\vspace{-1mm}
\section{Imitation Learning via Distribution Matching}
\vspace{-1mm}

\subsection{Problem Formulation}
\vspace{-1mm}
We use Markov Decision Processes (MDPs), a standard sequential decision-making model for our IL task. The MDP can be described as a tuple $\mathcal{M}=(\mathcal{S}, \mathcal{A}, P, \rho, \gamma)$, where $\mathcal{S}$ is the state space, $\mathcal{A}$ is the action space, $P\rbr{\cdot|s, a}: \mathcal{S} \times \mathcal{A} \rightarrow \Delta(\mathcal{S})$ is the transition operator with $\Delta(\mathcal{S})$ as the family of distributions over $\mathcal{S}, \rho \in \Delta(\mathcal{S})$ is the initial distribution and $\gamma \in(0,1)$ is the discount factor. 

The goal of offline IL is to find a policy $\pi: \mathcal{S} \rightarrow \Delta(\mathcal{A})$ that mimics the behavior of given expert demonstrations. In the offline IL setting, we cannot interact with the MDP environment to collect samples with the execution policy $\pi$, but only access a dataset of transitions sampled from the MDP, $\mathcal{D} = \{(s_i, a_i, s'_i),\mid (s, a)\sim q, s'\sim P(\cdot \mid s, a), i=1,2,\dots, N\}$, where $q$ is the data distribution. A subset $\Dcal^{\exp} = \{(s_i, a_i, s'_i),\mid (s, a)\sim d^{\exp}, s'\sim P(\cdot \mid s, a), i=1,2,\dots, N\} \subseteq \Dcal$ is the expert demonstrations, and $d^{\exp}$ is the distribution of state-action pairs generated by the expert. 
%

The IL can be completed by matching
the state-action stationary distribution $d^\pi(s)$ generated by execution policy~$\pi$
\begin{equation}
       \begin{aligned}
            d^\pi(s):=(1-\gamma) \sum_{t=0}^{\infty} \gamma^t \operatorname{Pr}(s_t=s\mid s_0 \sim \rho,\\
        a_t \sim \pi\left(s_t\right), s_{t+1} \sim P\left(\cdot\mid s_t, a_t\right),\forall t),
       \end{aligned}\label{eq:dpi_recursion}
        \end{equation} 
with the expert demonstration distributions $d^{\exp}$.  
We abuse the notation to denote the stationary discounted distribution on state-action pairs, $d^\pi\rbr{s, a}\defeq d^\pi\rbr{s}\pi\rbr{a\mid s} $. 



We formulate the distribution matching problem as minimizing the $f$-divergence between behavior state-action distribution $\dpi$ with expert demonstration distribution $d^{\exp}$ while satisfying the density constraints,
\begin{equation}
    \begin{aligned}
        \min_{\pi, \dpi} &~  D_{f}\rbr{\dpi\| d^{\exp }}   = \EE_{(s, a)\sim d^{\exp}} \sbr{f\rbr{\frac{\dpi(s, a)}{d^{\exp}(s, a)}}}  \\ 
        \st &~ \dpi\rbr{s', a'} = (1-\gamma) \rho\left(s'\right)\pi(a'| s')+\gamma\Pcal^\pi_* \dpi\rbr{s', a'}, \\
        &~ \forall \rbr{s', a'}\in\Scal\times\Acal.
    \end{aligned} \label{eq:original_opt}
\end{equation}
where $d$ is a general probability measure, $f$-divergence $D_f:\Delta(\Scal)\times\Delta(\Scal)\to \RR^+$ is a class of functions that measures the difference between two probability distributions defined with a \emph{convex} function $f:(0,\infty)\to\RR,f(1)=0$
and the $ \Pcal ^\pi _*$ transpose policy transition operator
\begin{equation}
    \mathcal P^\pi _* d(s', a') = \pi(a'\mid s')\iint P(s'\mid s, a)\dpi(s, a)dsda .\label{eq:p_transpose}
\end{equation}
The optimization~\eqref{eq:original_opt} is difficult to solve as the optimization variable is a function $\dpi\rbr{\cdot, \cdot}$ and there are infinite many constraints for each $\rbr{s, a}$ pair. Moreover, in offline setting, the execution policy $\pi$ is not able to interact with environments, increasing the optimization difficulty. 

\vspace{-1mm}
\subsection{Solution via Primal-dual Optimization}\label{sec.dice}
\vspace{-1mm}

We start by constructing and reformulation the Lagrangian of problem \eqref{eq:original_opt} with $Q(s, a)$ as the dual variable,

\vspace{-5pt}
{\small \begin{align}
&\max_{\pi, d} \min_Q -D_f\left(\dpi \| d^{\exp}\right)+\notag \\
& \iint Q(s, a) \cdot\left((1-\gamma) \rho(s) \pi(a | s)+\gamma \mathcal{P}_*^\pi \dpi(s, a)-\dpi(s, a)\right)dsda \label{eq:lag}  
\end{align}}
Noticing the problem~\eqref{eq:original_opt} is convex-concave given policy $\pi$, we can transform the Lagrangian \eqref{eq:lag} to
\begin{equation}
    \begin{aligned}
        & \max_{\pi}\min_Q~(1-\gamma) \cdot \mathbb{E}_{(s_0, a_0)\sim\rho\times\pi}\left[Q\left(s_0, a_0\right)\right]+\\
&\EE_{d^{\exp}} \sbr{\max_{\nu} \nu(s, a)\cdot\left(\gamma \mathcal{P}^\pi Q(s, a)-Q(s, a)\right) -f\rbr{\nu(s, a)} }
    \end{aligned}\label{eq:primal_dual}
\end{equation}
where we first reparametrize the primal variable $\dpi(s,a)$ to the density ratio $$\nu(s,a):=\frac{\dpi(s, a)}{d^{\exp}(s, a)}$$ and 
$\Pcal^\pi$ is the adjoint operator of $\Pcal^\pi_*$ defined as
$$
\mathcal P^\pi Q(s, a) = \iint P(s'\mid s, a)\pi(a'\mid s')Q(s, a)ds'da'. 
$$.
The detailed derivation is deferred to Appendix B in our online report \cite{online_report} due to space limit. 

\noindent\textbf{Simplification of \eqref{eq:primal_dual} via Fenchel duality.} Fenchel conjugate, or convex conjugate, indicates that any convex functions $f$ can be written as 
$$f(x)=\max_{\zeta\in\RR} \sbr{x\cdot\zeta-f^*(\zeta)}$$ for any $x\in(0,+\infty)$, where $f^*$ is the Fenchel conjugate or convex conjugate of $f$. $f^*$ is also a convex function. Then we can reformulate the second expectation in \eqref{eq:primal_dual} as 
\begin{equation}
    \begin{aligned}
    &\max_\nu \nu(s, a)\cdot\left(\gamma \cdot \mathcal{P}^\pi Q(s, a)-Q(s, a)\right) -f\rbr{\nu(s, a)} \\
    =&  f^*\rbr{\gamma \cdot \mathcal{P}^\pi Q(s, a)-Q(s, a)}
\end{aligned}\label{eq:convex_conj_optimality}
\end{equation}
Therefore, we can eliminate one optimization function and reduce the optimization complexity, resulting in 
\begin{equation}
    \begin{aligned}
    = &  \max_\pi\min_Q ~(1-\gamma)\cdot \mathbb{E}_{(s_0, a_0)\sim\rho\times\pi}\left[Q\left(s_0, a_0\right)\right] + \\
& ~\EE_{(s,a)\sim d^{\exp}}\sbr{f_*\rbr{\gamma\mathcal{P}^\pi\sbr{Q(s, a)}  - Q(s,a)}}
\end{aligned}\label{eq:valuedice}
\end{equation}
\textbf{Optimality conditions of $\nu$ and $Q$.} Most importantly, the optimal solution of convex conjugate in \eqref{eq:convex_conj_optimality} indicates the optimal primal $\nu^*_Q$ given dual variable $Q$ have the following relation,
\begin{equation}
     f'\rbr{\nu^*_Q(s, a)} = \gamma\mathcal{P}^\pi{Q(s, a)}  - Q(s,a)
    \label{eq:optimality}
\end{equation}
by computing the derivatives of LHS of \eqref{eq:convex_conj_optimality} and asking it to be zero. 

Moreover, when both primal variable $\nu$ and dual variable $Q$ are solved to the saddle points $\nu^*, Q^*$ given policy $\pi$, the constraints in \eqref{eq:original_opt} given $\pi$ are satisfied due to the saddle point optimality conditions. Therefore, the primal variable $d(s,a)$ recover $d^\pi(s,a)$ given $\pi$ in \eqref{eq:dpi_recursion}, indicating $\nu^*(s,a)=\frac{d^\pi(s,a)}{d^{\exp}(s,a)}$ and 
\begin{equation}
    f'(\nu^*(s,a))=f'\rbr{\frac{d^\pi(s,a)}{d^{\exp}(s,a)}} = \gamma\mathcal{P}^\pi{Q^*(s, a)}  - Q^*(s,a)
    \label{eq:optimality2}
\end{equation}
by extension of \eqref{eq:optimality}.
\begin{remark}[Interpretation of the dual variable $Q$.]\label{remark.interpretation}
Equation \eqref{eq:optimality} shares a similar formulation with the policy evaluation in RL, \ie, the dual variable $Q$ can be interpreted as the state-action value function with respect to the reward function $-f'\rbr{\nu^*_Q(s,a)}$. Moreover, when $Q$ is solved to optimal $Q^*$, it can be interpreted as the $Q$-function of $-f'\rbr{\frac{d^\pi(s,a)}{d^{\exp}(s,a)}}$. For example, if we select KL divergence where $f(x)=x\log (x)$, the reward function will be $-\log \frac{d^\pi(s, a)}{d^{\exp}(s, a)}  - 1$, \ie, the log density ratio. From now on, we will use KL divergence instead of general $f$-divergence, indicating the following relation (ignoring the constant shift),
\begin{equation}
     Q^*(s,a) = -\log \rbr{\frac{d^\pi(s,a)}{d^{\exp}(s,a)}} + \gamma\mathcal{P}^\pi{Q^*(s, a)} 
    \label{eq:q_def}
\end{equation}
\end{remark}



\begin{remark}[Difficulty in solving the $\min-\max$ optimization \eqref{eq:valuedice}]
    Solving \eqref{eq:valuedice} in practice is difficult in both computational and statistical aspects.  
\textbf{Statistically}, 
the imitation learning heavily relies on the expert trajectories in $d^{\exp}(s, a)$, which is expensive to collect. \textbf{Computationally}, with neural network introduced for parametrization of $Q\rbr{s, a}$, $\nu\rbr{s, a}$, and $\pi\rbr{a|s}$ in~\eqref{eq:valuedice}, the $\min$-$\max$ optimization is notoriously difficult and very unstable, therefore, usually requires many tuning and tricks, for example, gradient penalty from the generative adversarial training~\cite{salimans2016improved,kostrikov2019imitation}. These two factors together render inferior performance and poor generalization. 
\end{remark}



\section{Dynamics Representations}

In this section, we define the \emph{dynamics representations} from the factorization of system dynamics. We show that the dynamics representations can significantly help IL since they can fully represent $Q(s, a)$ in \eqref{eq:convex_conj_optimality}, enabling us to constrain optimization in the representation space. Moreover, the dynamics representations can be learned from arbitrary data sharing the same transition dynamics, reducing the statistical dependency on expert data only.


\subsection{Definitions}

\begin{definition}[Dynamics representations]\label{def.repr}
     There exists representations ${\phi}: \mathcal{S} \times \mathcal{A} \rightarrow \mathbb{R}^k$ and ${\mu}: \mathcal{S} \rightarrow \mathbb{R}^k$ such that
     \begin{equation}
         P(s'\mid s, a ) = \inner{\phi(s, a)}{\mu(s')p_n(s')} \label{eq:factorization}
     \end{equation}
     where the $p_n\in\Delta(\Scal)$ is a noise distribution that has full support on $\Scal$, and $\inner{\cdot}{\cdot}$ is vector inner product. 
     \qed
\end{definition}
\begin{remark}[Noise distributions $p_n$ and connection to linear MDP\cite{jin_provably_2019,agarwal_flambe_2020,ren2022spectral}.] Our factorization is similar to the linear MDP literature in the RL theory community except for the additional noise term $p_n(s')$ inspired by \cite{zhang2022making}. Adding the extra noise term has two benefits: \textbf{a)} Aligning with the density ratio learning in the offline setting when setting $p_n$ as $d^{\rm exp}$ shown in Section \ref{sec.repr_capacity};
\textbf{b)} Enabling tractable loss function to learn representations $\phi,\mu$ shown in Section \ref{sec:practical_rl}. 
\end{remark}

\begin{remark}[Transferability and connections to successor features~\cite{dayan1993improving}.] We emphasize that our dynamics representations are only relevant with dynamics $P$, which can be naturally transferred across data collected from different policies or tasks sharing the same dynamics. Another popular family of feature transfer leverages the successor features \cite{dayan1993improving,barreto2017successor} sharing the similar decomposition of $Q(s,a)=\inner{\psi^\pi(s, a)}{w}$, which is obtained from the factorization of reward functions $r=\inner{\mathbf{r}(s, a)}{w}$ and $\psi^\pi(s, a) = \EE_{\pi}[\mathbf{r}(s, a)]$. Note that the representation $\psi^\pi$ is relevant with policy $\pi$, constraining its reuse within similar tasks only. Our dynamics representations $\phi,\mu$ are irrelevant to policy, indicating more general transferability. Moreover, for our IL problem, no reward functions are explicitly defined, making leveraging the successor representations not practical.

\end{remark}

Some might question the existence of such factorizations. We show an example of stochastic control with known dynamics,
\begin{example}[Fournier random feature for nonlinear control with Gaussian noise \cite{ren2023stochastic}.] \label{example}
    We give an example of features for a known nonlinear control system with Gaussian noise that is commonly seen in robotic control, \ie 
    $$
    s'=g(s, a) + \epsilon,\quad \epsilon\sim\Ncal(0,\sigma^2I_d)
    $$
    where $g:\Scal\times\Acal \to\Scal$ is a deterministic nonlinear dynamics, and $\epsilon$ is the Gaussian noise. We can regard the transition dynamics as 
    $P(s'\mid s, a) \propto \exp(-\|g(s, a)-s'\|^2/2\sigma^2)$, which is a Gaussian kernel. Then, according to Bonchner's theorem \cite{devinatz1953integral}, we have the Fourier random features of the Gaussian kernel,
    $
     P(s'\mid s, a) = \inner{\psi_\omega(g(s, a))}{\psi_\omega\left(s^{\prime}\right)}_{\Ncal(\omega)}
    $
    where $\psi_w(x)=\exp(\mathrm{i}\omega^\top x)$ and $\inner{\cdot}{\cdot}_{\Ncal(\omega)}= \mathbb{E}_{\omega \sim \mathcal{N}\left(0, \sigma^{-2} I_d\right)}\left[\inner{\cdot}{\cdot} \right]$. It translated into infinite-dimensional representations $\phi,\mu$ whose $i^{\rm th}$ element are $$\phi_i(s, a)=\psi_{w_i}(g(s, a)), \mu_i(s')=\psi_{w_i}(s')/p_n(s')$$, respectively, where $\omega_i\sim\Ncal(0,\sigma^{-2}I_d)$. In this case, $\phi,\mu$ are both infinite-dimensional features, but we can use finite-dimensional truncation as an approximation in practice with provable approximation guarantees~\cite{ren2023stochastic}.
\end{example}

    
\subsection{Representational Capacity of Dynamics Representations}\label{sec.repr_capacity}
We show that the proposed dynamics representations $\phi, \mu$ can fully represent the dual variable $Q$ in \eqref{eq:convex_conj_optimality}. We first show what is the representation space of $\phi,\mu$, respectively.

\noindent\textbf{Density ratio $\frac{d^\pi(s)}{p_n(s)}$ represented by $\mu(s)$.} Recalling the recursion on $d^\pi(s)$ in \eqref{eq:original_opt}, substituting the factorization \eqref{eq:factorization} in it, and dividing both sides by $p_n(s')$, we have
\begin{equation}
    \begin{aligned}
    \frac{d^\pi(s')}{p_n\rbr{s'}} = &\rbr{1 - \gamma}\frac{\rho(s')}{p_n\rbr{s'}}\\
    + &\gamma \inner{\mu\rbr{s'}}{\underbrace{\int \phi\rbr{s, a}{d^\pi(s)} \pi\rbr{a|s}ds da }_{\theta^\pi}}.\label{eq:linear_nu_s}
\end{aligned}
\end{equation}
where $\theta^\pi$ is a linear weight irrelevant with $s'$. As we know the initial distribution $\rho$ and the noise distribution $p_n$, we have the full linear representations of $\mu(s)$, or the state density ratio $\frac{d^\pi(s)}{p_n(s)}$. Moreover, when setting $p_n=d^{\exp}$, the state-action density ratio can be further represented by
\begin{equation}
    \begin{aligned}
    &\nu^*(s,a) = \frac{d(s)}{d^{\exp}(s)}\frac{\pi(a\mid s)}{\pi^{\exp}(a\mid s)} \\
     &:=\rbr{\rbr{1 - \gamma}\frac{\rho(s)}{p_n\rbr{s}} + \gamma \inner{\mu\rbr{s}}{\theta^\pi}}\zeta(s, a)
\end{aligned}\label{eq:dsa_factorization}
\end{equation}
where $\zeta(s, a):=\frac{\pi(a\mid s)}{\pi^{\exp}(a\mid s)}$ is the policy ratio.

\noindent\textbf{Dual variable $Q$ represented by $\phi,\mu$ jointly}. From the optimal solution \eqref{eq:optimality2} and Remark \ref{remark.interpretation} we observe that the optimal dual variable $Q^*$ can be interpreted as the value function of reward $-\log\rbr{\nu^*(s,a)}$. Substitute the dynamics representations \eqref{eq:factorization} into \eqref{eq:optimality}, 
\begin{equation}
    \begin{aligned}
        & Q^*(s, a) =  -\log \nu^*(s, a)+ \gamma\Pcal^\pi Q^*(s, a) \\
    = & -\log \nu^*(s, a)  + \\
    & \inner{\phi(s, a)}{\underbrace{\gamma\int\mu(s')p_n(s')\pi(a'| s')Q(s', a')ds'da'}_{\omega^{\pi}}}
    \end{aligned} 
\end{equation}
Substitute the density factorization \eqref{eq:dsa_factorization},
 we have the full representation of optimal dual variable $Q^*$,
\begin{equation}
    \begin{aligned}
        &\quad \quad Q^*_{w^\pi\!,\theta^\pi}(s, a)=\\
        &-\underbrace{\log\zeta(s, a)}_{\substack{\text{Offline}\\\text{computable}}} + \underbrace{ \phi^\top(s, a)\omega^{\pi}\!\!-\!\log \!\Big({\mu(s)^\top\theta^{\pi}} }_{\text{Sit in the representation space}}\!\!+\!\underbrace{(1-\gamma)\frac{\rho(s)}{d^{\exp}(s)}}_{\text{Offline computable}}\!\!\Big)
    \end{aligned}\label{eq:linar_repr_q}
\end{equation}
where $\omega^\pi\in\RR^k$ is the linear weights on representations $\phi$ independent from $s, a$. 
Note that the policy ratio $\zeta$ and initial state density ratio $\frac{\rho(s)}{d^{\exp}(s)}$ are all offline computable from the expert dataset and current behavior policy. Therefore, we have the full parametrization structure of $Q^*$, where the parameters to optimize are only $\omega^\pi,\theta^\pi$, \ie, the coefficients on representations $\phi, \mu$, respectively. In practice, we can directly put parameters $\omega,\theta$ and optimize with gradient descent.


\subsection{Imitation Learning with Dynamics Representations}
As we know that the representational structure of optimal $Q^*$, we can constrain the optimization of dual variable $Q$ within the representation space according to \eqref{eq:linar_repr_q}, 
%
we can substitute the representation of $Q^*$  
to the our objective function \eqref{eq:valuedice} and transferring optimizing $Q$ to optimizing the parameters $\omega,\theta$, \ie,
\begin{equation}
\begin{aligned}
\max_\pi\min_{\omega,\theta} & \underset{(s, a) \sim d^{\exp }}{\mathbb{E}}\sbr{\exp\rbr{ \gamma\mathcal{P}^\pi{Q^*_{\omega,\theta}(s, a)}  - Q^*_{\omega,\theta}(s,a)}} \\
+ &(1-\gamma) \underset{\substack{(s_0, a_0) \sim \rho\times\pi}}{\mathbb{E}}\left[Q^*_{\omega,\theta}(s_0, a_0)\right].
\end{aligned}
\label{eq:il_dual_valuedice}
\end{equation}
where $f_*(x)=\exp(x)-1$ when we select KL divergence (constants in the objective functions are ignored).
With the learned representation, we can constrain the \texttt{min} side of the optimization to the representation space, making it easier to solve \texttt{min-max} problems.



\vspace{-1mm}
\begin{figure*}[ht]
    \centering
    \includegraphics[width=\linewidth]{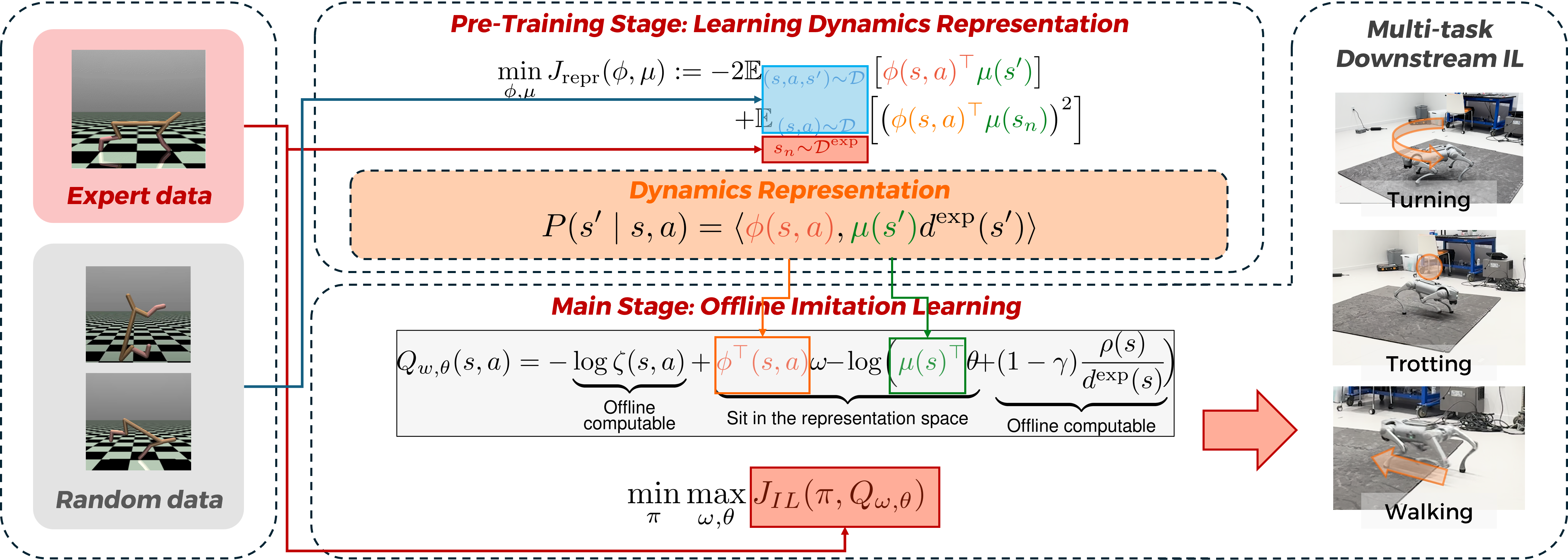}
    \caption{Demonstration of the proposed algorithm. We first learn dynamics representations in the pre-training stage, and learn downstream IL in the main training stage within the representation space.}
    \label{fig:demo}
\end{figure*}
\section{Two-stage Algorithm for Representation and Imitation Learning}
\vspace{-1mm}

In this section, we propose a two-stage algorithm and discuss practical considerations. We will first explain a tractable representation learning algorithm as the pre-training stage that admits arbitrary datasets sharing the same dynamics $P$, then discuss the main training stage conducting downstream IL on representation space.
\subsection{Pre-training Stage: Dynamics Representation Learning}\label{sec:practical_rl}
We show how to learn the representations $\phi,\mu$ from data. We consider the case where we have a (small) expert dataset $\Dcal^{\exp}$ and a (large) general dataset $\Dcal$ containing all data generated from the same dynamics $P$. Then we define the following representation learning objective function,

\begin{equation}
    \begin{aligned}
        &\min_{\phi, \mu} J_{\rm repr}(\phi,\mu):= -2\EE_{(s,a, s')\sim \Dcal}\sbr{\phi(s,a)^\top \mu(s')}\\
        ~&\quad\quad\quad\quad\quad\quad\quad+ \EE_{(s,a) \sim \Dcal, s_n \sim \Dcal^{\exp}}\sbr{(\phi(s,a)^\top \mu(s_n))^2}\label{eq:repr_learning}
    \end{aligned}
\end{equation}
We show that by minimizing \eqref{eq:repr_learning} we can get representation satisfying factorizations in \eqref{def.repr} since 
\begin{equation}
\begin{aligned}
&J_{\rm repr}(\phi,\mu)= \EE _{(s,a)\sim\Dcal} \sbr{\int l_{\phi,\mu}(s,a,s')ds'} -C ~\text{where} \label{eq:repr_squaure_form}\\
    &~ l_{\phi,\mu}(s,a,s')\!:=\!\left(\!\frac{P(s' \mid s,a)}{\sqrt{p_n(s')}}\! -\! \phi(s,a)^\top \!\mu(s') \sqrt{p_n(s')}  \!\right)^2
\end{aligned}
\end{equation}
$C$ is a constant irrelevant with $\phi$ and $\mu$. The detailed derivation can be found in Appendix B in our online report \cite{online_report}. 
It is easy to see from \eqref{eq:repr_squaure_form} that the $\phi$ and $\mu$ minimizing $J_{\rm repr}$ satisfy the factorization in \eqref{eq:factorization}.
 To improve numerical stability, we add a log probability regularization term similar to \cite{zhang2022making},
\begin{equation}
    \min_{\phi,\mu}\! J_{\rm repr}(\phi,\mu)\! +\! \lambda_{\rm repr} \EE_{(s, a)\sim \Dcal}\!\sbr{\rbr{\log \rbr{\EE_{s'}\!\sbr{\phi^\top(s, a)\mu(s')}}}^2}\label{eq:practical_repr_learning}
\end{equation}
where $J_{\rm repr}(\phi,\mu) $ is defined in \eqref{eq:repr_learning}, and $\lambda_{\rm repr}$ is the regularization weights. 


\subsection{Main Stage: IL on Representation Space}\label{sec:practical_IL}

Equation \eqref{eq:il_dual_valuedice} already showed the IL on representation space, we address some practical considerations here. In practice, the initial states can be sampled from $d^{\exp}$ without affecting the optimality (See discussion in Section 5.3 in \cite{kostrikov2019imitation}). Therefore, the $\frac{\rho(s)}{d^{\exp} (s)}$ equals constant $1$. Moreover, to avoid learning log policy ratio $\log\zeta(s,a)$ offline, we directly use a neural network $f_\xi(s, a)$ to fit it and arrives at practical $Q$ parametrization,
\begin{equation}
    Q_{\omega,\theta,\xi}(s,a) = f_\xi(s, a) + \phi(s, a)^\top\omega -\log\rbr{\mu(s)^\top\theta + 1 - \gamma}
    \label{eq:q_params}
\end{equation}
We solve the inner dual variable estimation and outer policy optimization alternatively.

The overall two-stage IL algorithm is presented in Algorithm \ref{alg:abstract} as well as in Figure \ref{fig:demo}.  
\begin{algorithm}[H]
    \caption{Representation ValueDICE (Abstract version)}
    \label{alg:abstract}
    \begin{algorithmic}[1]
        \REQUIRE Expert trajectory dataset $\Dcal^{\exp}$, Dynamics dataset $\Dcal\supseteq \Dcal^{\exp}$,  initial policy $\pi_0$
        \STATE \texttt{\textcolor{blue}{\# representation learning stage}}
        \STATE Solve for representation $\phi,\mu$ via solving \eqref{eq:practical_repr_learning} with dynamics dataset $\Dcal$ and expert dataset $\Dcal^{\exp}$.
        \STATE \texttt{\textcolor{blue}{\# Imitation learning stage, alternatively do dual variable evaluation and policy update.}}
        \WHILE{Training, alternatively}
        \STATE Parameterize $Q$ with learned representation $\phi,\mu$ and parameters $\omega,\theta,\xi$ following \eqref{eq:q_params}.
        \STATE Optimize $\omega,\theta,\xi$ parameters of dual variable $Q$  via 
        optimization \eqref{eq:il_dual_valuedice}.
        \STATE Update policy $\pi$ via 
        optimization \eqref{eq:il_dual_valuedice}.
        \ENDWHILE
    \end{algorithmic}
\end{algorithm}

\section{Experiments}

The experimental tests aim to justify our claims on the representation-based DICE imitation learning, \ie, 
\begin{itemize}
    \item \textit{Does the representation improve IL performance and generalizations when expert data is limited?}
    \item \textit{Can we reuse the representation from data sharing the same dynamics to help imitation learning?}
\end{itemize}
We will show results on locomotion tasks in MuJoCo simulators (with expert data only and a combination of expert and random data) and real quadrupeds. 
\subsection{Imitation Learning with Limited Expert Data}\label{exp:mujoco}

\begin{figure}[h]
    \centering
    \includegraphics[width=0.2\linewidth]{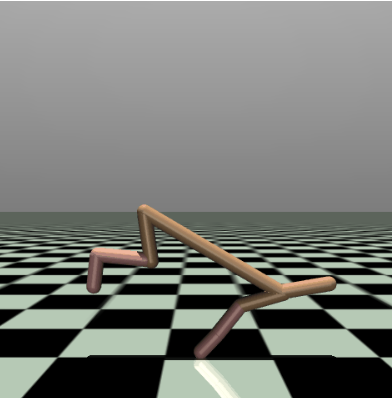}
    \includegraphics[width=0.2\linewidth]{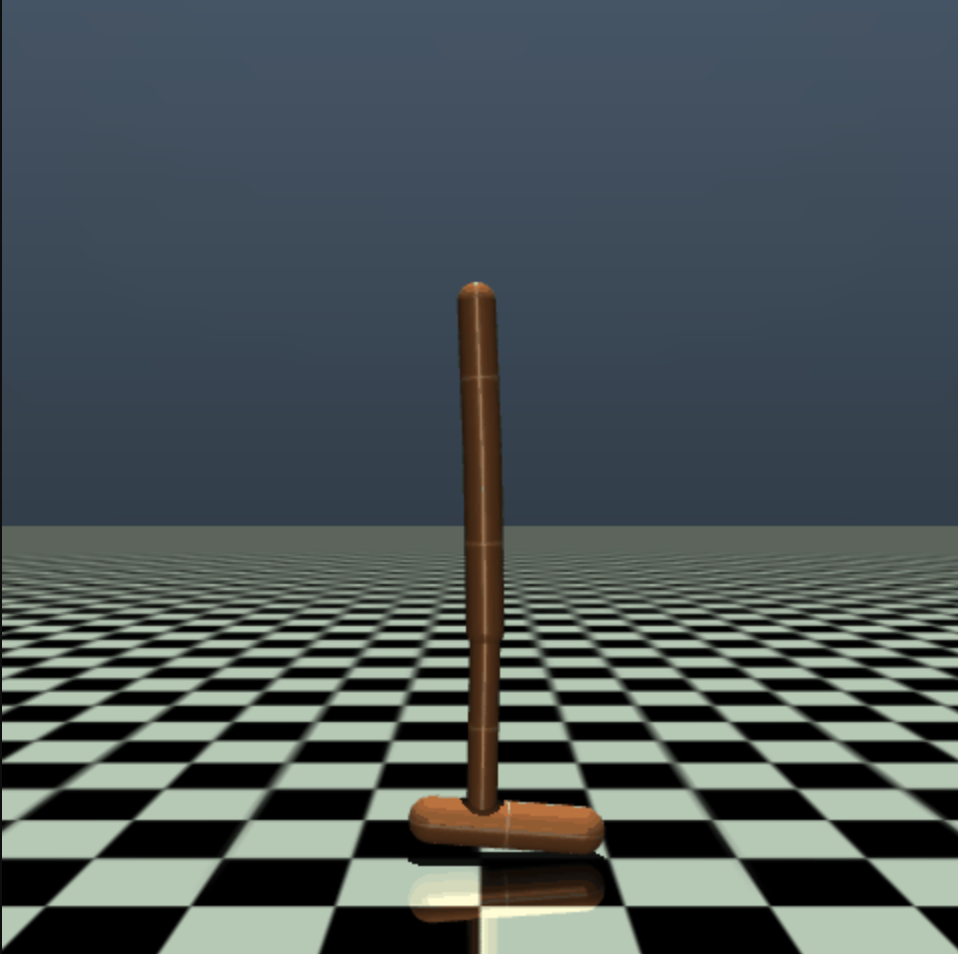}
    \includegraphics[width=0.2\linewidth]{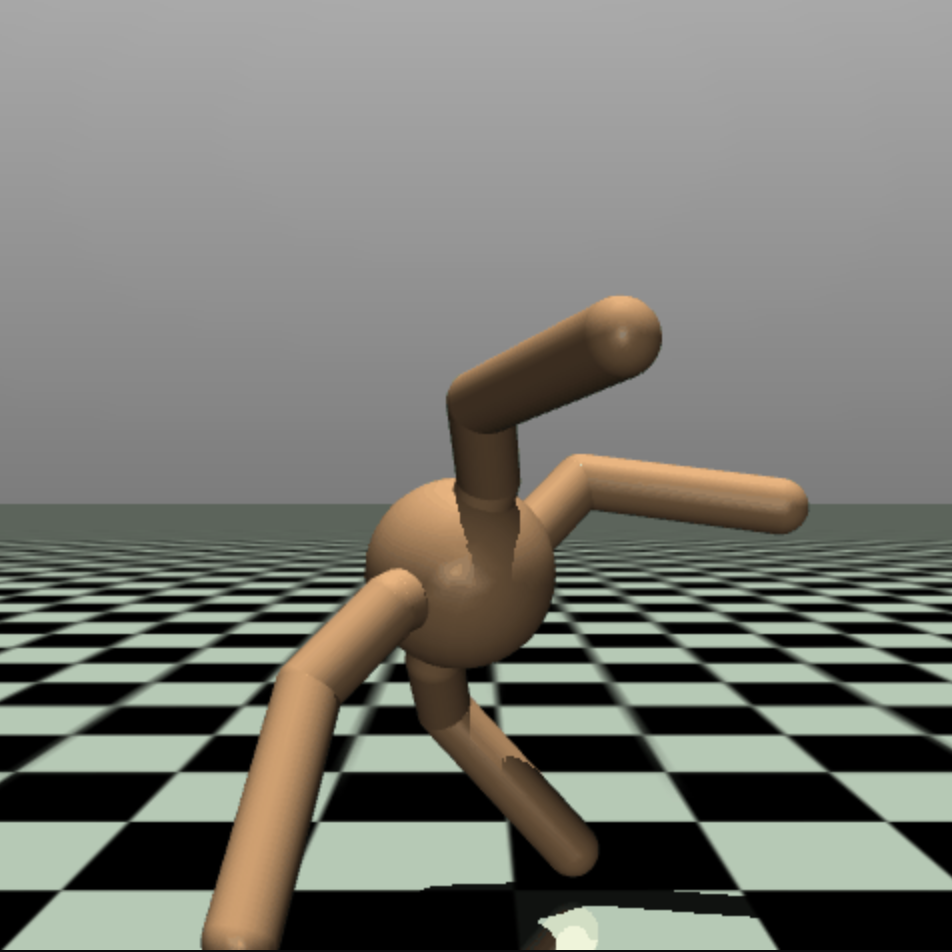}
    \includegraphics[width=0.2\linewidth]{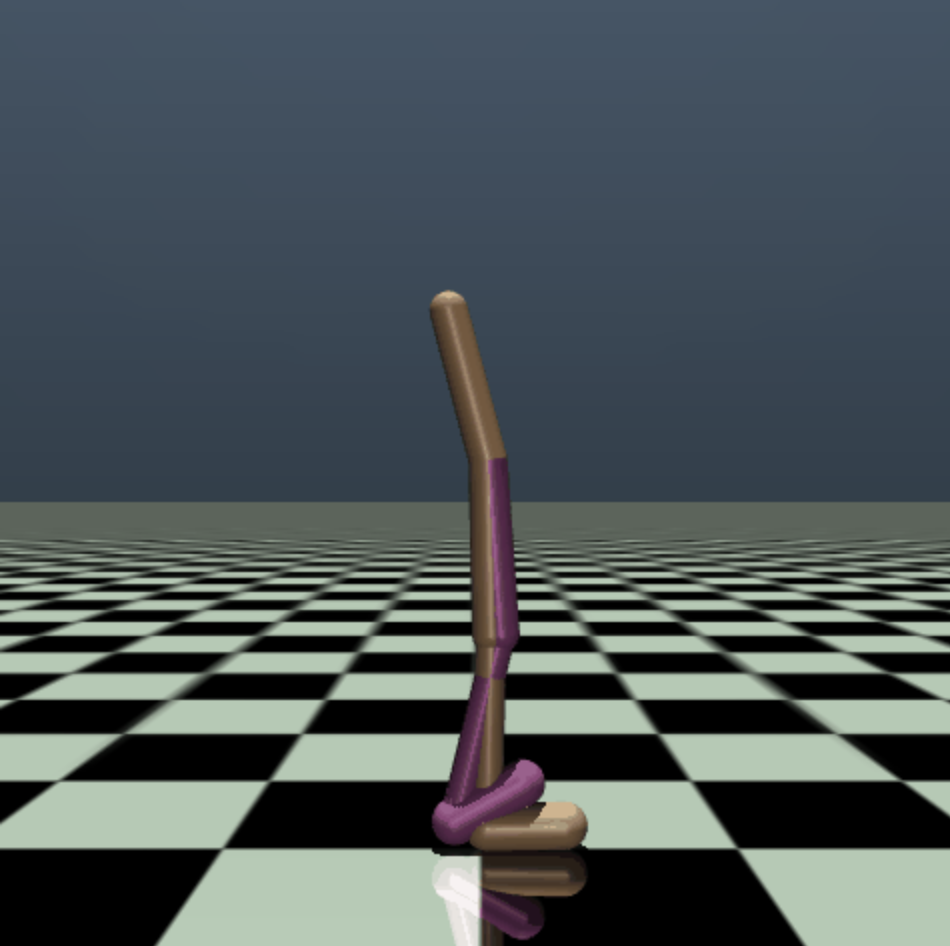}
    
    \caption{Locomotion tasks in the MuJoCo simulator, from left to right: Half Cheetah, Hopper, Ant, and Walker2d.}
    \label{fig:mujoco}
\end{figure}

\noindent\textbf{Tasks and expert data.} We conduct imitation learning on four locomotion tasks using MuJoCo physics simulator \cite{todorov2012mujoco} shown in Figure \ref{fig:mujoco}, which are commonly used in previous IL papers \cite{kostrikov2019imitation,ho2016generative}. To answer question 1, we sample a limited number of expert trajectories from the expert dataset provided in \cite{kostrikov2019imitation}, \ie, \textbf{1 trajectory} for HalfCheetah, Hopper, and Walker2d, \textbf{3 trajectories} for Ant. The original ValueDICE paper \cite{kostrikov2019imitation} provides 40 trajectories. Each trajectory has a total of 1000 transitions. 


\noindent\textbf{Algorithms and Baselines.} Other than the proposed algorithm \textbf{ReprValueDICE}, We include baselines (1) \textbf{ValueDICE} directly uses a NN to parameterize dual variable $Q$ \cite{kostrikov2019imitation} and (2) behavior cloning (\textbf{BC}) that leverages maximum likelihood estimation to match expert and behaviour policies.

\begin{figure}[h]
    \centering
    \includegraphics[width=0.4\linewidth]{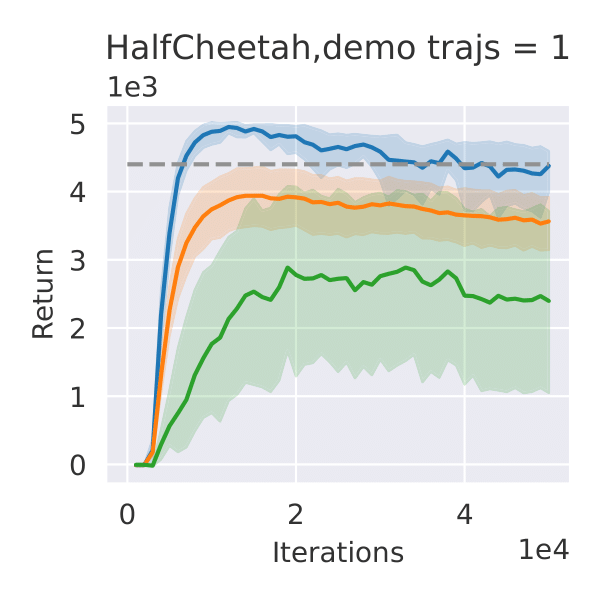}
    \includegraphics[width=0.4\linewidth]{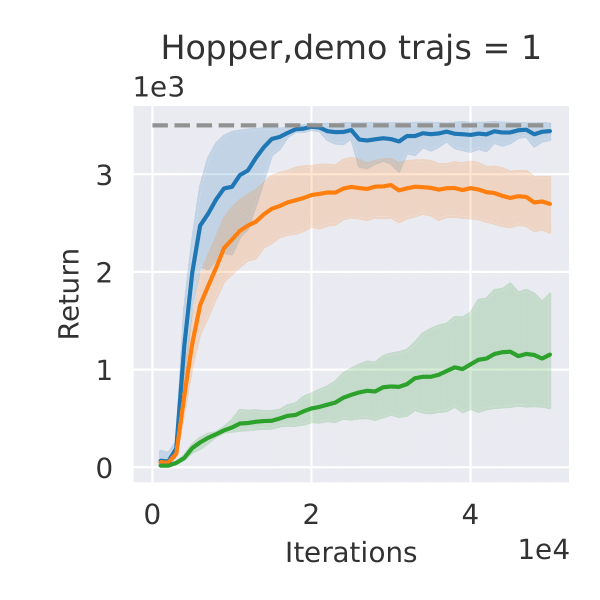}\\
    \includegraphics[width=0.4\linewidth]{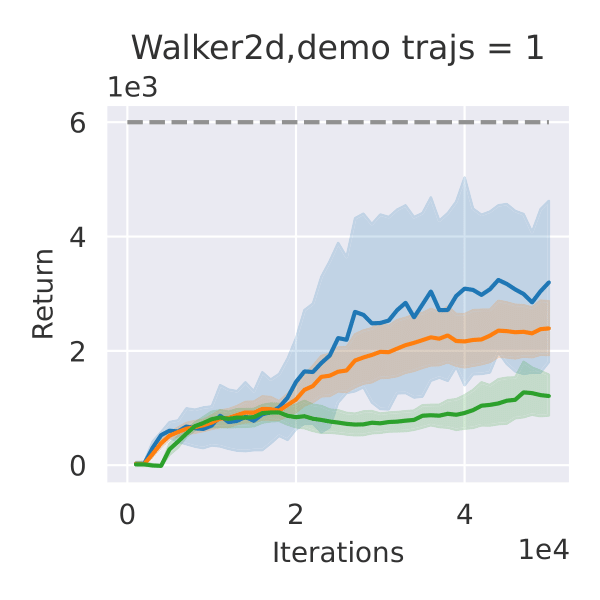}
    \includegraphics[width=0.4\linewidth]{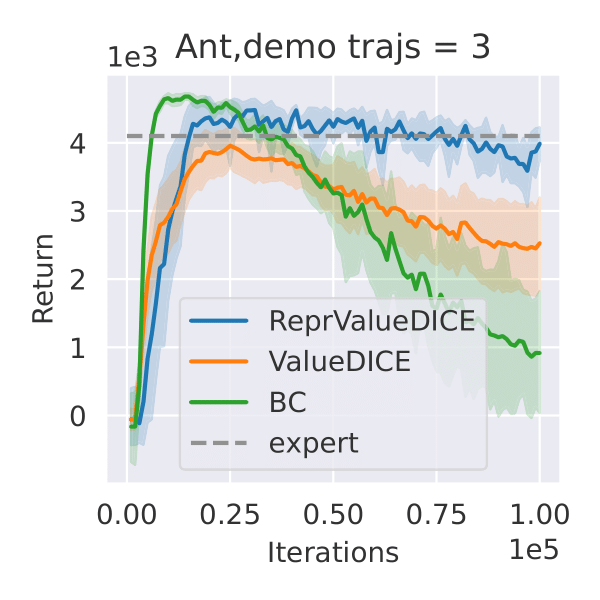}
    \caption{Training performance of learning from limited expert data. Solid lines and shaded regions show the mean and standard deviation of episodic returns per 1000 training iterations with 10 random seeds.}
    \vspace{-2mm}
    \label{fig:exp1}
\end{figure}

\noindent\textbf{Evaluation and Performance.} We evaluate the IL policy over 20 randomly initialized trajectories to assess the performance and demonstrate the generalization capabilities of policies learned from only 1 or 3 trajectories only, shown in  
Figure \ref{fig:exp1}. The results show that ReprValueDICE achieves significantly better average returns than ValueDICE on HalfCheetah, Hopper, and Walker2d. For the Ant, ReprValueDICE shows stable performance consistently over the whole training process towards the expert, while BC and ValueDICE show good performance initially but then deteriorate quickly. All the results have shown that the dynamics representation can help mitigate overfitting and improve performance even with a little expert data, verifying that exploiting dynamics representation can improve the IL performance and generalization to unseen trajectories. 

\begin{figure}[h]
    \centering
    \includegraphics[width=0.4\linewidth]{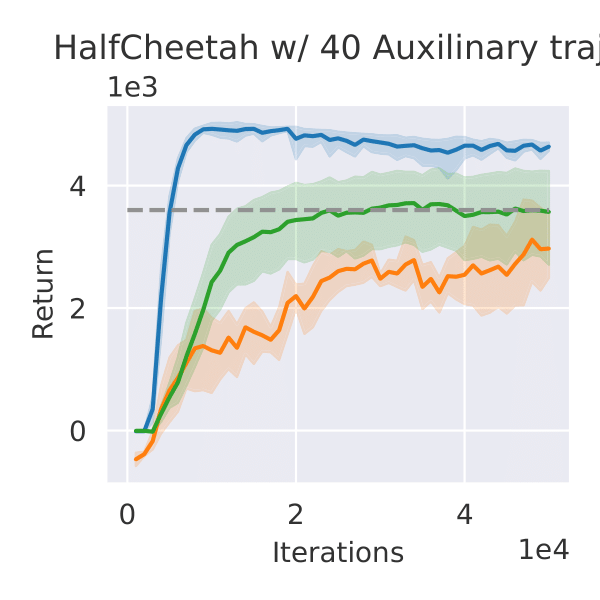}
    \includegraphics[width=0.4\linewidth]{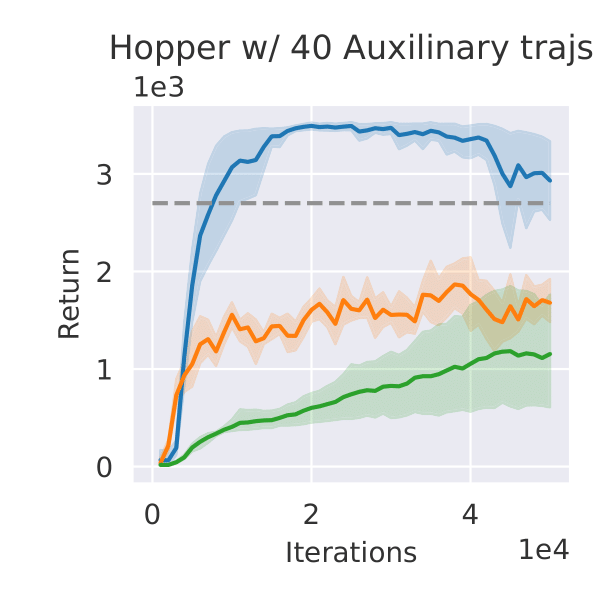}\\
    \includegraphics[width=0.4\linewidth]{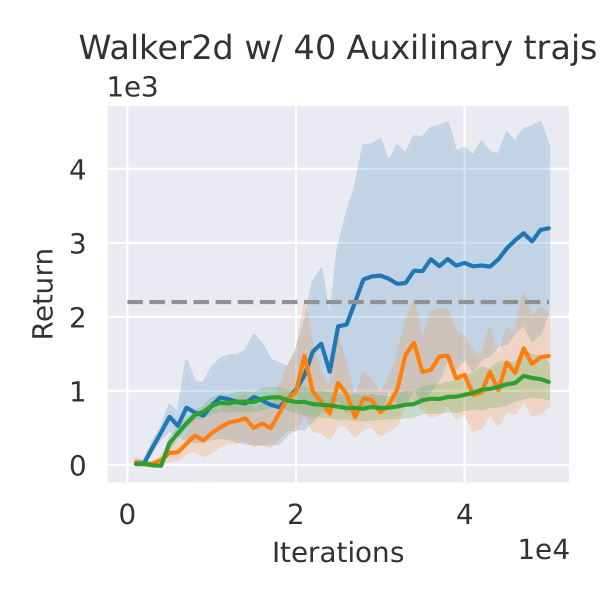}
    \includegraphics[width=0.4\linewidth]{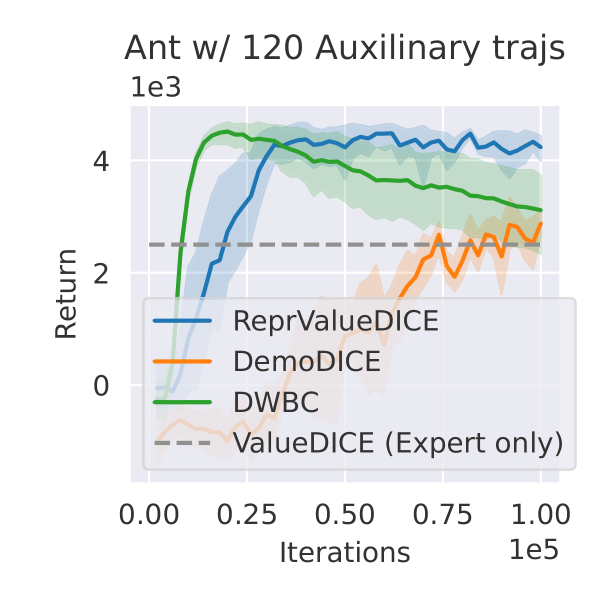}
    \caption{Training performance of learning from expert data and auxiliary data. Solid lines and shaded regions show the mean and standard deviation of episodic returns per 1000 training iterations with 10 random seeds. Dashlines indicate learning the performance learning from expert data only using ValueDICE~\cite{kostrikov2019imitation}.}
    \label{fig:exp2}
\end{figure}
%


For sensitivities to other hyperparameters like the number of extra random data, the number of expert demonstrations, please refer to Appendix D in our online reports \cite{online_report}.

\subsection{Imitation Learning with Auxiliary Data}

\noindent\textbf{Experimental Setup.} We continue to conduct IL on MuJoCo locomotion tasks and leverage auxiliary data from D4RL~\cite{fu2020d4rl} medium-level datasets to further help imitation learning. We consider appending 40 times more trajectories (120 trajectories for Ant and 40 trajectories for the other 3) of the previously mentioned mujoco tasks. 

\noindent\textbf{Algorithm and baselines.} Our \textbf{ReprValueDICE} leverages the auxiliary data ane expert data together to learn representations in the pre-training stage and then imitates the expert data in the main stage. We compare our ReprValueDICE against a) \textbf{DemoDICE}~\cite{kim2022demodice} that uses the suboptimal data as regularization and 2) discriminator-weighted behavior cloning (\textbf{DWBC})~\cite{xu2022discriminator}.

\noindent\textbf{Evaluation and performance.} We use the same evaluation process as Section \ref{exp:mujoco}. Experimental results have shown that ReprValueDICE shows significant improvements over ValueDICE with expert data only and outperforms DWBC and DemoDICE. Results on DWBC and DemoDICE show that when the medium-level data dominates the dataset, where in our case \texttt{expert data}:\texttt{medium data}=$1:40$, handling auxiliary data via regularization or weighted BC cannot achieve good performance.


\subsection{Imitating Real Quadrupeds Controllers}

We further show the superior expert data efficiency of proposed representation-based imitation learning on the real Unitree Go2 Quadrupeds trying to imitate the built-in controller on a task of walking, \ie, tracking given base speed commands. Specifically, we consider three tasks, trotting (zero speed), walking ($0.5{\rm m/s}$ linear velocity) and turning ($0.5~{\rm rad/s}$ angular velocity).
\begin{figure}[H]
    \centering
    \includegraphics[height=0.3\linewidth]{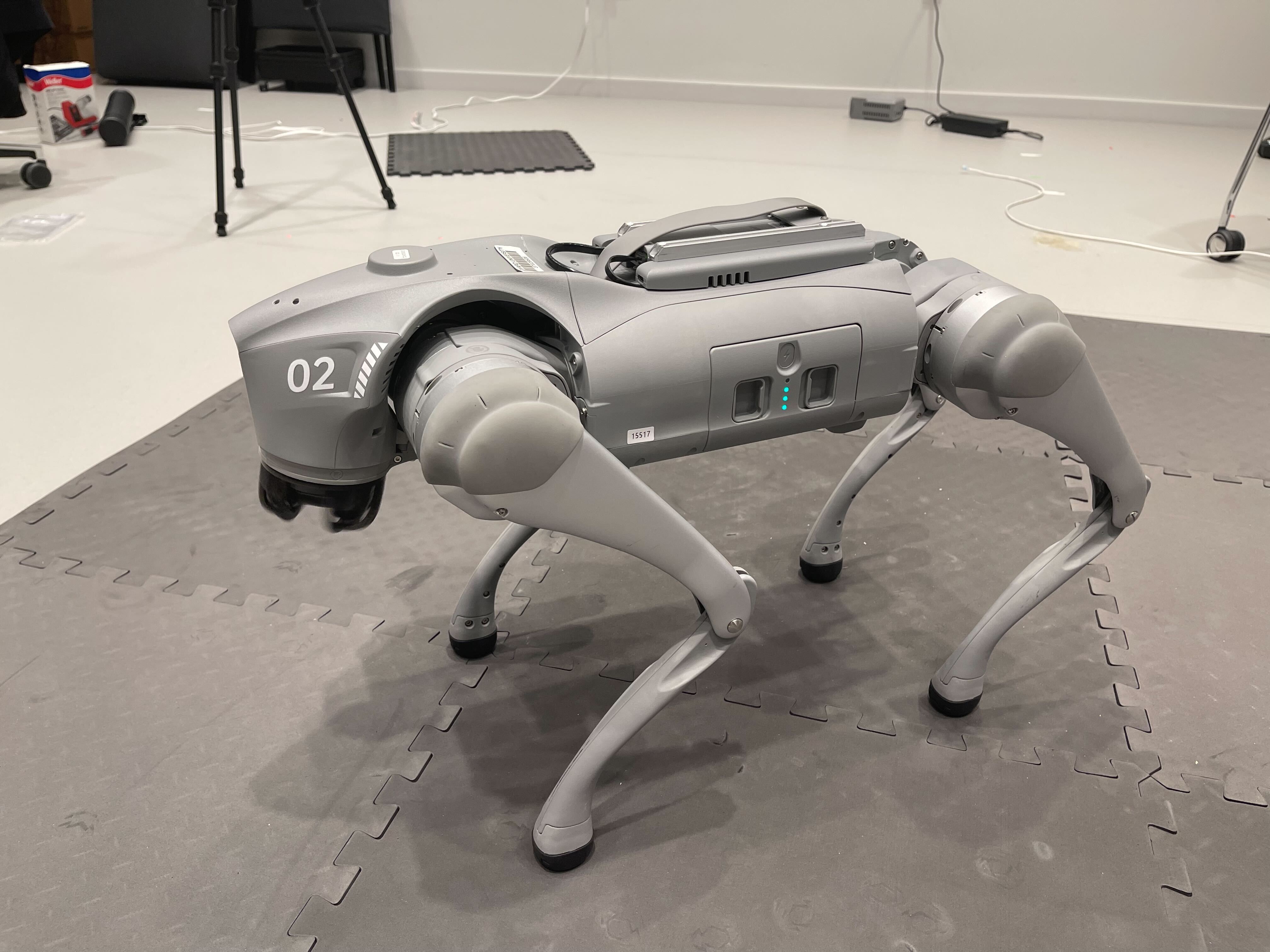}
    \includegraphics[height=0.3\linewidth]{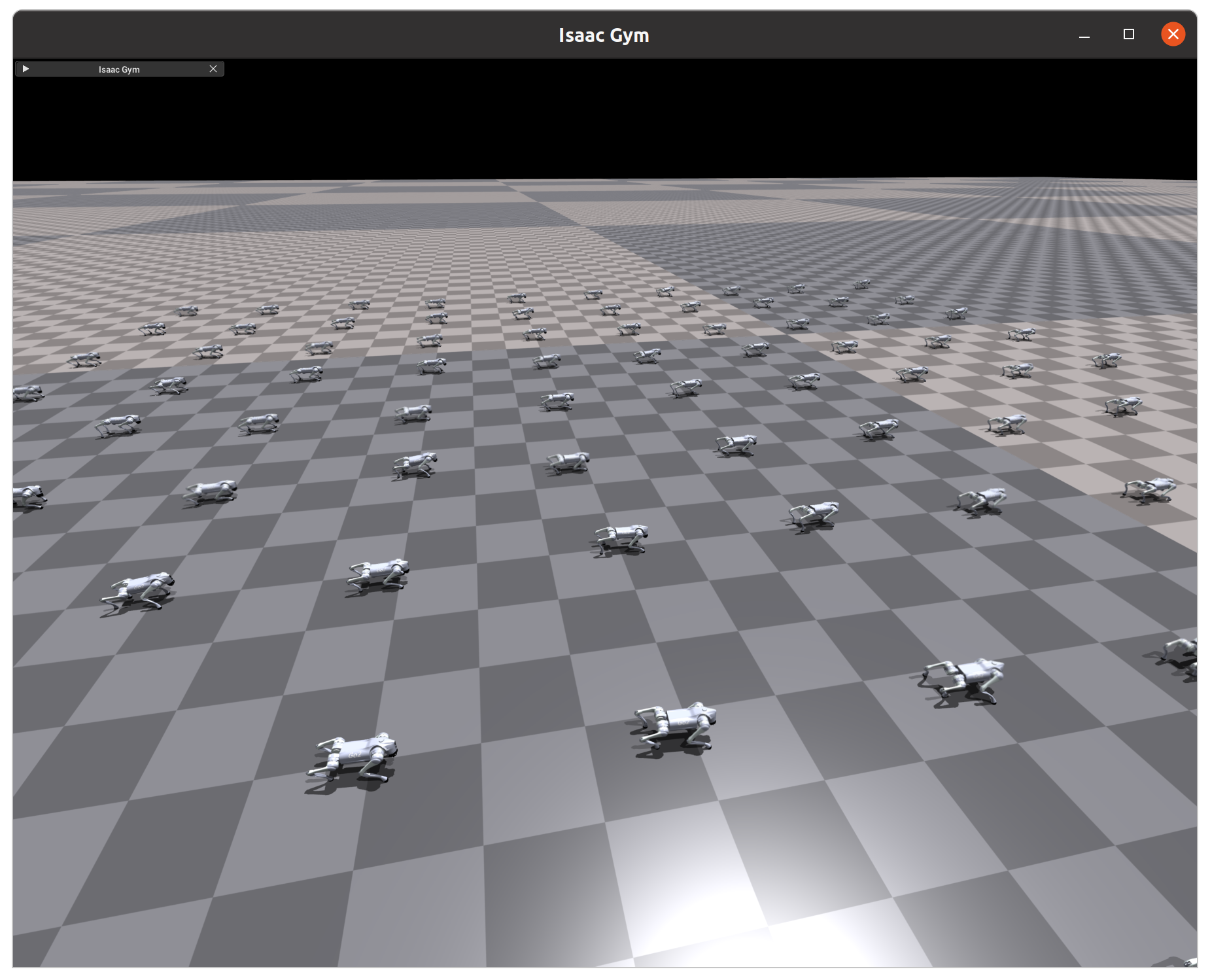}
    \caption{ We use the Unitree Go2 Quadrupeds and control it via the Jetson Orin NX extension deck (left). Auxiliary data are collected from the Issac Gym simulator (right).}
    \label{fig:go2}
\end{figure}

\noindent\textbf{Real-world and Simulator Data Collection.}  We collect 1000 seconds of real-world expert data using the built-in controller. Meanwhile, to show that we can learn and reuse representations from arbitrary data, we also use the Isaac Gym simulator to collect a large amount of auxiliary data. The data is collected during the process of training a reinforcement learning (RL) agent and is used as the dynamics dataset $\Dcal$ to learn representations in equation \eqref{sec:practical_rl}.

\noindent \textbf{Baselines and metrics.} We compare the proposed ReprValueDICE with ValueDICE and BC that learns from expert data collected from real world only, showing that by leveraging dynamics representation, we can learn an effective walking controller with only a small amount of data. To get quantitative metrics on the gait behaviors, we compute the base walking height, foot air percentage, and average contact force partitions of front foot.

\noindent \textbf{Results.} We record the gait behavior metrics in Table \ref{tab.perf_real}, which shows that the proposed ReprValueDICE well mimics the stock controller and demonstrates appropriate base height, less air time, and balance contact (average 25\% on the front foot), all show a stable walking behavior. The RL always leans forward with a higher base, which easily leads to instability under external perturbation. As for ValueDICE and BC, they all fail to stably walk after learning from the same amount of data. Moreover, the velocities tracking performance of the real quadrupeds are shown in Appendix D in our online report~\cite{online_report}, showing the successful imitation of stock controllers. The video of learned policy are can also be found on our \href{https://congharvard.github.io/repr-imitation-learning/}{Project webpage}.

\begin{table}[h]
{\small
\centering
\caption{Gait behaviour comparison averaging over all three speed tracking tasks. \textbf{The closer to the target stock controller in {\color{blue}blue}, the better the algorithm is}. Behavior cloning (BC) and ValueDICE from expert data only failed to walk stably. }\label{tab.perf_real}
\resizebox{\columnwidth}{!}{
\begin{tabular}{@{}rccc@{}}
\toprule
                 & \begin{tabular}[c]{@{}c@{}}Base walking \\ height (cm)\\ \end{tabular} & \begin{tabular}[c]{@{}c@{}}Foot air time \\ percentage$^*$(\%)\\ \end{tabular} & \begin{tabular}[c]{@{}c@{}}Average front foot \\ contact force partition \\ \end{tabular} \\ \midrule
Stock (Target Policy) &         {\color{blue}{32.4}{\small $\pm$2.2}  }      & {\color{blue}{24.0}{\small $\pm$3.3} }                     &        {\color{blue}{24.8}{\small $\pm$12.2}   }               \\\midrule
ReprValueDICE   &          \textbf{31.6}{\small $\pm$4.9}           &  \textbf{25.0}{\small $\pm$8.5}                     &      \textbf{27.6}{\small $\pm$14.5}                     \\
RL    &         {37.8}{\small $\pm$5.3}            &  {30.0}{\small $\pm$11.3}                    &       {42.4}{\small $\pm$14.5}                  \\ 
BC    &              N/A      &    N/A            &           N/A         \\
\begin{tabular}[r]{@{}r@{}}ValueDICE \\ (from expert demo)\\ \end{tabular}   &              N/A      &    N/A            &           N/A         \\
\bottomrule
\end{tabular}
}}
\end{table}


\vspace{-1mm}
\section{Concluding Remarks}
\vspace{-1mm}

In this paper, we propose the dynamics representations to address the challenge of optimization and sample efficiency in offline IL. We define representations through a factorization of transition dynamics and show that it can fully represent the decision variable $Q$ in offline IL. Experimental results on MuJoCo and real quadrupeds verified that the proposed algorithm has less overfitting and better generalization and can reuse auxiliary non-expert data to learn representations and improve algorithm performance. 

\bibliographystyle{imitation_learning/format/IEEEtran}
\bibliography{imitation_learning}
\clearpage
\appendices
\end{document}